\documentclass{llncs}
\usepackage{xspace}
\usepackage{url}
\usepackage{graphicx}
\usepackage{xspace}
\usepackage{color}
\newcommand{\mtwo}{M$^{2}$\xspace}
\newcommand{\protege}{Prot\'eg\'e\xspace}
\newcommand{\toolname}{Populous\xspace}

\newcommand{\partof}{\emph{part\_of}}

\newcommand{\comment}[2]{}

\newcommand{\vocab}[1]{{\texttt{#1}}}

\usepackage{makeidx}  
\begin{document}
\frontmatter          

\title{Populous: A tool for populating Templates for OWL ontologies}
\author{Simon Jupp\inst{1} \and Matthew Horridge\inst{1} \and Luigi Iannone\inst{1} \and Julie Klein\inst{2}
\and Stuart Owen\inst{1}  \and Joost Schanstra\inst{2} \and Robert Stevens\inst{1} \and Katy Wolstencroft\inst{1}}

\institute{School of Computer Science, University of Manchester, Manchester, UK
\and
Institut National de la Sant\'e et de la Recherche M\'edicale, Toulouse, France}

\maketitle

\makeatother
\begin{abstract}
We present  \toolname, a tool for gathering content with which to populate an ontology.
Domain experts need to add content, that is often repetitive in its form, but without having to tackle the underlying ontological representation.
\toolname presents users with a table based form in which columns are constrained to take values from particular ontologies; the user can select a concept from an ontology via its meaningful label to give a value for a given entity attribute.
Populated tables are mapped to patterns that can then be used to automatically generate the ontology's content.
\toolname's contribution is in the knowledge gathering stage of ontology development.
It separates knowledge gathering from the conceptualisation and also separates the user from the standard ontology authoring environments. As a result, \toolname  can allow knowledge to  be gathered in a straight-forward manner that can then be used to do mass production of ontology content.
 \end{abstract}


\section{Introduction}
\label{sec:intro}

Ontologies are being developed to provide controlled vocabularies for the annotation of life science data. Annotating data with ontologies adds semantics to the data that can facilitate data integration and enrich data analysis\cite{citeulike:1882392}, \cite{1427291}, \cite{citeulike:212874}. For ontologies to have a faithful representation of a domain, experts from that domain must have input to the authoring process. There are barriers that prevent domain experts engaging in an ontology's development; in particular, the semantics of an ontology language, or the intricacies of the authoring tools. To address this issue we have developed \toolname that allows users to contribute their knowledge by populating simple templates that are then transformed to statements in the underlying representation language. 

When designing an ontology it is often the case that repeating patterns occur in the modelling. These patterns can be abstracted from the ontology and used to specify simple templates that could be populated by domain experts~\cite{gangemiISWC2005,ekaw2008,bmc_odps}. As an example of a pattern, consider an ontology about cells; eukaryotic cells can be classified as being either anucleate, mono-nucleate, binucleate or multinucleate. We can abstract over this pattern to say that every cell can be classified by its nucleation. This pattern is repeated for all cell types; the only variables are the cell name and the value for its nucleation. We can now use this pattern to build a simple template that could be populated by a cytologist, without him or her ever knowing about the underlying ontological representation.  This type of pattern is common in ontology development where you have one set of entities being described in terms of another set of entities~\cite{rector}. 

The tabular layout provides a simple and intuitive form fill-in style of user interface for a user to populate such templates. Typically, each row corresponds to a set of related entities and each column represents the type of relationship.  The intersection of row and column holds the `filler' for the given entity's relationship of that column's type. By adopting templates ontology developers can separate the pattern from its population; this allows the domain expert to focus on the knowledge without the distraction of the knowledge representation language.

Templates are useful when data needs to be collected in a regular form. Applying constraints to the template reduces the amount of discrepancies in the input data. A common tool for collecting data in this form is the spreadsheet; spreadsheets provide a tabular interface, where columns and rows represent certain attributes, and individual cells capture the data. Tables help users to structure data in a logical way, that is useful for both its maintenance and processing.  In ontology development spreadsheets can be used to gather and organise information about concepts and their relationships. Previous work in this area has focused on the transformation of data into ontologies, but little attention has been paid to supporting the population of the templates at the point of data entry and this is where \toolname's main contribution lies. 

\subsection{Previous work}

Various tools are available to support the conversion of spreadsheet data into statements in a knowledge representation language. Excel2RDF\footnote{\url{http://www.mindswap.org/~rreck/excel2rdf.shtml}}, Convert2RDF\footnote{\url{http://www.mindawap.org/~mhgoeve/convert/
}} and RDF123~\cite{rdf123} are three tools that allow users to generate Resource Description Framework (RDF) statements from spreadsheets. Despite RDF being the reference syntax for the Web Ontology Language (OWL), its serialisation is complex and not intended for humans, making it inappropriate for defining higher level OWL construct in patterns. 

The ExcelImporter plugin\footnote{\url{http://protegewiki.stanford.edu/wiki/Excel_Import}} for \protege 4.0 was a step up from these tools and enabled users to transform spreadsheets content directly into OWL axioms. It was, however, limited to only a small set of OWL constructs. The more recent tools to support template data and pattern instantiation include Mapping Master~\cite{m2}, OPPL~2~\cite{eswc09-luigi,ekaw2008} and the \protege Matrix plugin\footnote{\url{http://protegewiki.stanford.edu/wiki/Matrix}}:

\begin{itemize}
\item{The MappingMaster plugin for the \protege 3.4 ontology editor is a more flexible tool for transforming arbitrary spreadsheet data into OWL. MappingMaster moves away from the row centric view of spreadsheets and has an expressive macro language called \mtwo~\cite{m2_2,m2} that can handle non-uniform and complex spreadsheets.  \mtwo combines a macro language for referring to cells in a spreadsheet with a human readable syntax for generating OWL expressions called the Manchester OWL Syntax~\cite{mos}. MappingMaster and \mtwo are primarily designed for the transformation of spreadsheet data to OWL, but provides little in the way of support for populating and validating the spreadsheet data.}
\item{The Ontology Pre-Processing Language (OPPL)~\cite{ekaw2008,eswc09-luigi} (version 2) is a scripting language similar to \mtwo.  OPPL 2 is also Manchester OWL Syntax based and allows for the manipulation of OWL ontologies at the axiom level. OPPL 2 has support for the use of variables and the addition and removal of logical axioms from an ontology. OPPL 2 is a powerful scripting language for OWL and a user interface is provided via the OPPL plugin for \protege 4.0. OPPL, however, does not currently support working with tabular data and is decoupled from any knowledge gathering.}
\item{The MatrixPlugin for \protege 4.0 allows users to specify simple OWL patterns in a tabular interface that can be used to populate repeating patterns with existing concepts from an ontology. This plugin is useful for ontology developers that have repetitive patterns to instantiate, and has the added benefit of cell validation and auto-completion at the point of data entry. The Matrix plugin is limited by the type of patterns that can be expressed along with the fact that it is tightly integrated with the \protege interface, therefore, not suitable for all users. It does, however, combine knowledge gathering and axiom generation.} 
\end{itemize}

\subsection{Requirements}

All of the previous tools developed in this area tend to focus on the transformation from the template to the ontology. They provide little or no support for populating and validating template content. Furthermore, tools like ExcelImporter, OPPL and MappingMaster are integrated into the ontology development tool, so they are aimed at users that are already familiar with ontology development.  A table based tool for ontology authoring should shield the user from the underlying ontology and help guide the user when populating the template. Providing validation at the time of authorship should significantly reduce the amount of time required to debug and process the data captured in the spreadsheet. Here we list some key requirements for an ontology based table editor:

\begin{enumerate}
\item{Concepts may be described in terms of other concepts from other ontologies. In setting up \toolname the users must be able to load and browse ontologies.}
\item{The contents in the column of a table need to be restricted to concepts from imported ontologies, or parts of imported ontologies.}
\item{To improve human comprehension the concept should be rendered using only the URI fragment, or optionally a human readable label from the ontology.}
\item{Each time a concept is added to a cell within the table \toolname needs to check that the concept is valid according to the constraints resulting from requirement 2.}
\item{A cell might have multiple values; for example, when the concept being described has multiple parts.}
\item{Users should be free to suggest new concepts when an appropriate concept is not available.}
\end{enumerate}

\section{\toolname}

\toolname is an extension of RightField\cite{microsoft}; RightField is for creating Excel documents that contain ontology based restrictions on spreadsheets content. RightField enables users to upload Excel spreadsheets, along with ontologies from their local file systems, or from the BioPortal~\cite{citeulike:4716987} (a repository of biological ontologies available at \url{http://bioportal.bioontology.org}). RightField supports OWL, OBO and RDFS ontologies. Using RightField, individual cells, or whole columns or rows can be marked with the required ranges of ontology terms. For example, they could include all subclasses from a chosen class, direct subclasses only, all individuals, or only direct individuals. Each spreadsheet can be annotated with terms from multiple ontologies. RightField is primarily designed for generating spreadsheet templates for data annotation; \toolname extends RightField to support knowledge gathering and ontology generation. \toolname and RightField are both open source cross platform Java application. They use the Apache-POI~\footnote{\url{http://poi.apache.org}} for interacting with Microsoft documents and manipulating Excel spreadsheets. Populous is available for download from here \url{http://www.e-lico.eu/populous}.

Requirement 1 is already addressed using RightField functionality to upload both OWL and OBO ontologies. In order to better serve the life science community, users can also browse and load ontologies directly from BioPortal. Once the ontologies are loaded they are classified by a reasoner and the basic class hierarchy can be viewed. 

Requirement 2 is met by the ability to select terms from the ontology to create validation sets. A data validation restricts the set of values that are valid for a particular cell in the table. Validations can span multiple rows and columns and be composed of classes, properties or individuals from the ontology.  These data validations are stored in hidden worksheets along with additional information such as the full URI for the term, a label and the source ontology URI. These templates can also be exported as Microsoft Excel documents, which include the data validations on cells. 

We address requirement 3 by allowing users to populate cells using ontology labels. Once data has been entered the default will be to render the ontology term using its label; if no label is specified the URI fragment is used.  RightField already supports reading Microsoft Excel workbooks so users are free to populate the templates in external tools before importing them into \toolname for validation and transformation. 

By using \toolname directly users will benefit from having instant validation of the input data, satisfying requirement 4, along with some advanced features such as regular expression based auto-completion as they type into annotated cells. Additionally \toolname supports the addition of multiple values into a single cell that are validated individually according to requirement 5.  This can be particularly useful for certain kinds of patterns where a conjunction of variables is required to construct the axiom (see Section \ref{sec:usecase} for example). \toolname also allows the addition of free text values, even if the cell has an associated validation range, thus satisfying requirement 6. These values are highlighted to the user in red and can act as placeholders for new or suggested terms when no suitable candidate could be found in the validation set. 

\toolname supports the use of OPPL patterns in order to generate new OWL axioms from the populated template. OPPL scripts can be written directly in \toolname's design mode  or imported from scripts generated in the OPPL plugin. Variables from the OPPL pattern must be mapped to columns from the table using the column name. A pattern Wizard guides the user through the generation and execution of the OPPL scripts. When the template is processed new identifiers for unknown terms can be auto-generated and exported from \toolname. 

\begin{figure}
\centering
\includegraphics[height=7cm]{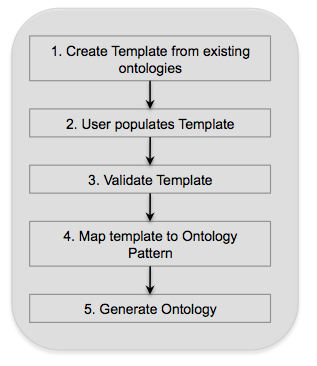}
\caption{\label{fig:workflow}Populous workflow}
\end{figure}
\subsection{Building an Ontology with \toolname}

\begin{figure}
\centering
\includegraphics[height=8.5cm]{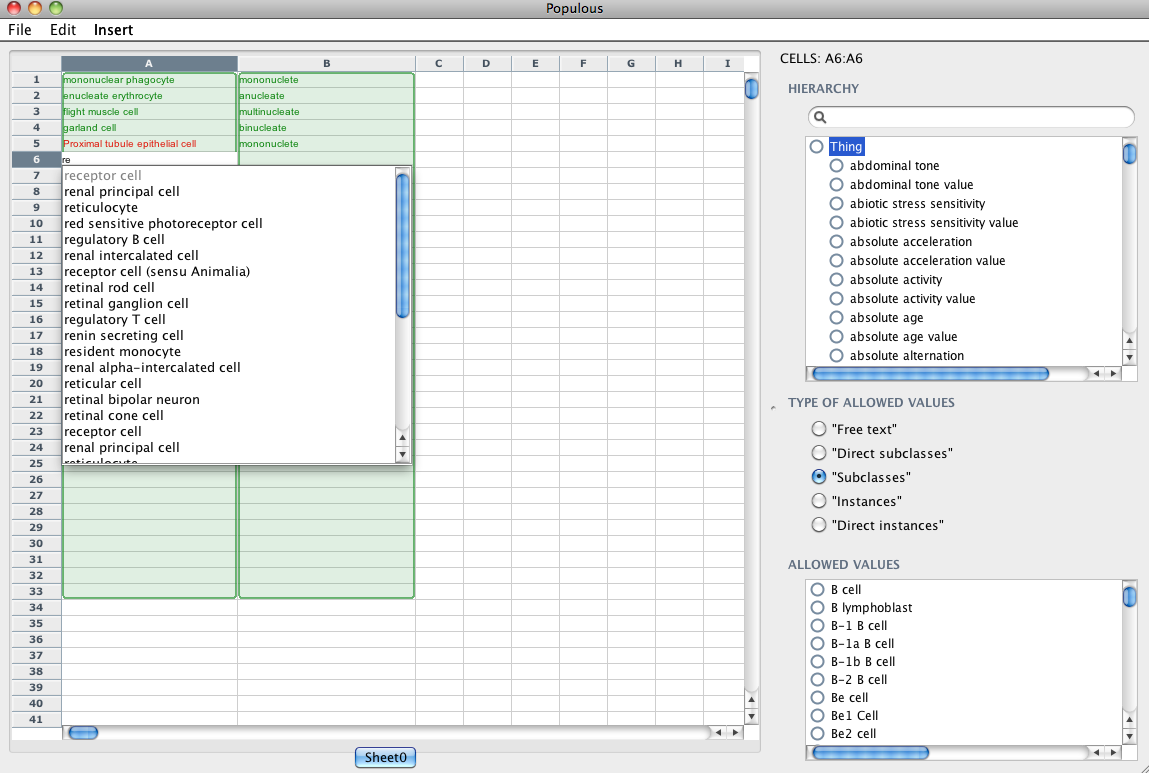}
\caption{\label{fig:populous_cell}Screenshot of Populous showing template population for cell types and nucleation}
\end{figure}

\begin{figure}
\centering
\includegraphics[height=7cm]{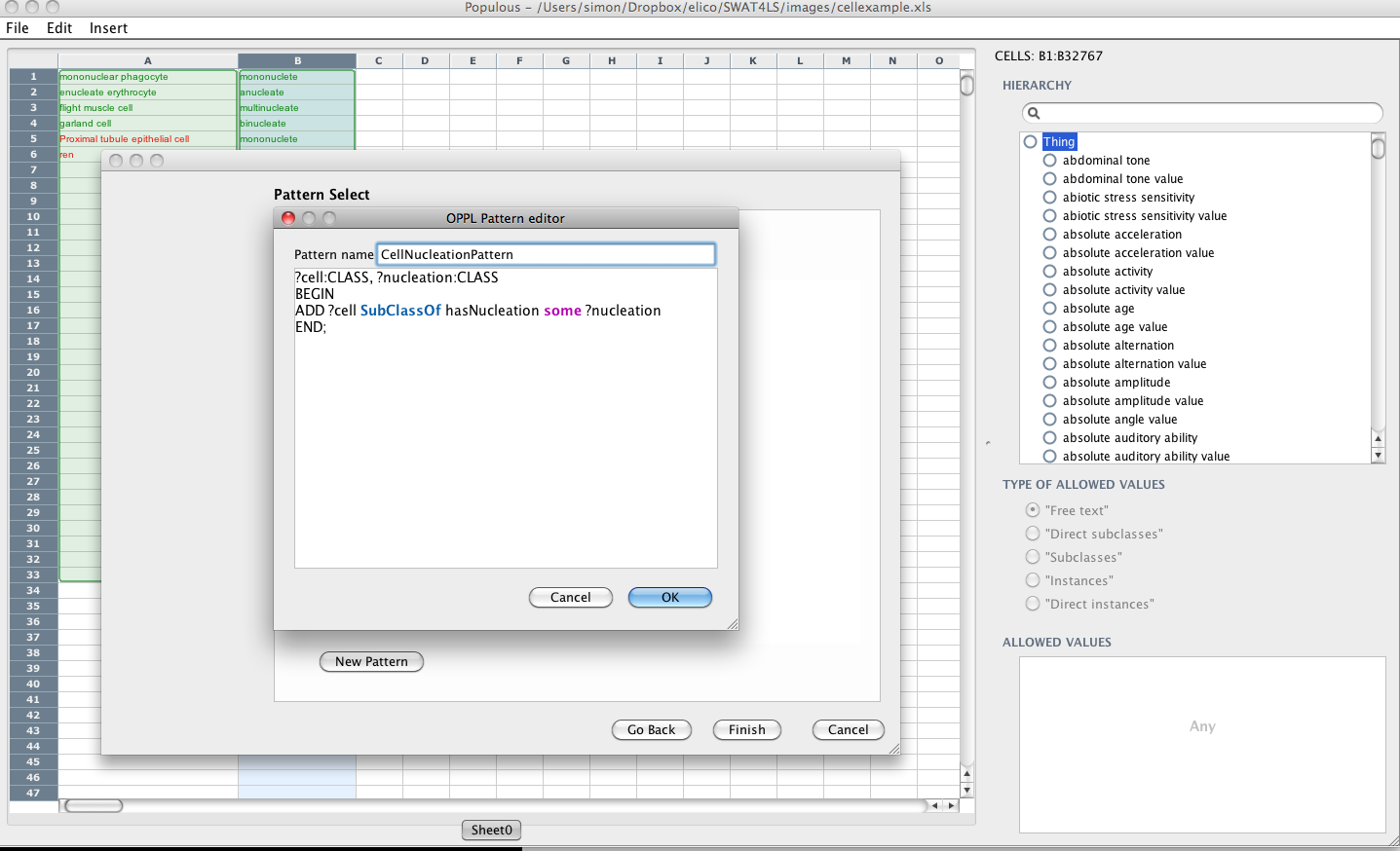}
\caption{\label{fig:populous_wizzard}Screenshot of Populous Pattern Wizard showing the OPPL script editor}
\end{figure}

We can demonstrate \toolname in building a simple ontology about cell types as described in Section \ref{sec:intro}. The pattern in the ontology states that every cell must have a nucleation. We need to create a template with two columns, column A is for cell type concepts, whilst column B is for nucleation concepts. Ontologies describing cells and their nucleation already exist that we can import into \toolname. By connecting to BioPortal we can load the Cell Type Ontology (CTO)~\cite{cto} and Phenotype and Trait Ontology (PATO)~\cite{citeulike:566073}. In order to restrict column A to terms from the CTO, we highlight all the cells in column A and restrict them to all subclasses of the root class. Column B is restricted subclasses of the \emph{nucleation} concept from PATO. The template is now ready to be populated by the domain expert. 

Figure \ref{fig:populous_cell} shows a part populated template. The terms in green indicate a valid term has been entered into the cell. The term in Column A5, \emph{Proximal tubule epithelial cell} is red because it is not a valid term from the CTO. Cell A6 is in the process of being edited with the auto-completer offering valid suggestion for input. 

The populated spreadsheet can now be transformed into an ontology. This can be done using the pattern wizard in \toolname (Figure \ref{fig:populous_wizzard}). The first step in the pattern wizard asks the user to select the columns and rows that contain populated data. In this example the pattern creates a restriction on each cell stating that all cells have a relationship, called \emph{hasNucleation}, to an instance of the class nucleation. This pattern can be expressed in OPPL 2 with the syntax shown in Figure \ref{fig:oppl}.

\begin{figure}
\vocab{?cell:CLASS,}\\
\vocab{?nucleation:CLASS}\\
\vocab{BEGIN}\\
\vocab{	ADD ?cell SubClassOf hasNucleation some ?nucleation}\\
\vocab{END;}\\
\caption{\label{fig:oppl} OPPL 2 pattern for cells and nucleation}
\end{figure}

There are two variables in the pattern, \texttt{?cell} and \texttt{?nucleation}. These variable are mapped to column A and B respectively. The pattern is to be instantiated using data from rows one to six that must be specified in the Wizard. The next step involves validating the pattern, given that \emph{Proximal tubule epithelial cell } is unknown by the validator, the user is given the option to assign a new URI for this concept. The final step generated the full OPPL script for applying this pattern. The Manchester OWL syntax generated for row one is shown in Figure \ref{fig:mos_cell}. A complete grammar for the OPPL 2 syntax is available here\footnote{\url{http://oppl2.sourceforge.net}}.

\begin{figure}
\vocab{Class: cto:CL\_0000113}\\
\vocab{SubClassOf:}\\
\vocab{hasNucleation some pato:PATO\_0001407}\\
\caption{\label{fig:mos_cell} Mononuclear Phagocyte described in Manchester OWL syntax generated form the OPPL 2 pattern in figure \ref{fig:oppl}  (PATO\_0001407 is the identifier for \emph{mononucleate})}
\end{figure}
 
\section{Use case and evaluation}
\label{sec:usecase}

In order to evaluate \toolname in a real ontology building scenario, it has been used to populate a template for gathering knowledge about the kidney and urinary system. The kidney is a complex organ composed of several distinct anatomical compartments that together enable the filtration of waste from the blood in the form of urine. Each of the kidney compartments is formed from a wide variety of cell types, and the specificity of the compartments relies on these specialised cell functions. The Kidney and Urinary Pathway Ontology (KUPO)~\cite{kupo} describes kidney cells, their function and their anatomical locations. KUPO is being built to annotate and integrate a variety of KUP related data held in the Kidney and Urinary Pathway Knowledge Base (KUPKB)\footnote{\url{http://www.e-lico.eu/kupkb}}.  
 
\begin{figure}
\centering
\includegraphics[height=8.5cm]{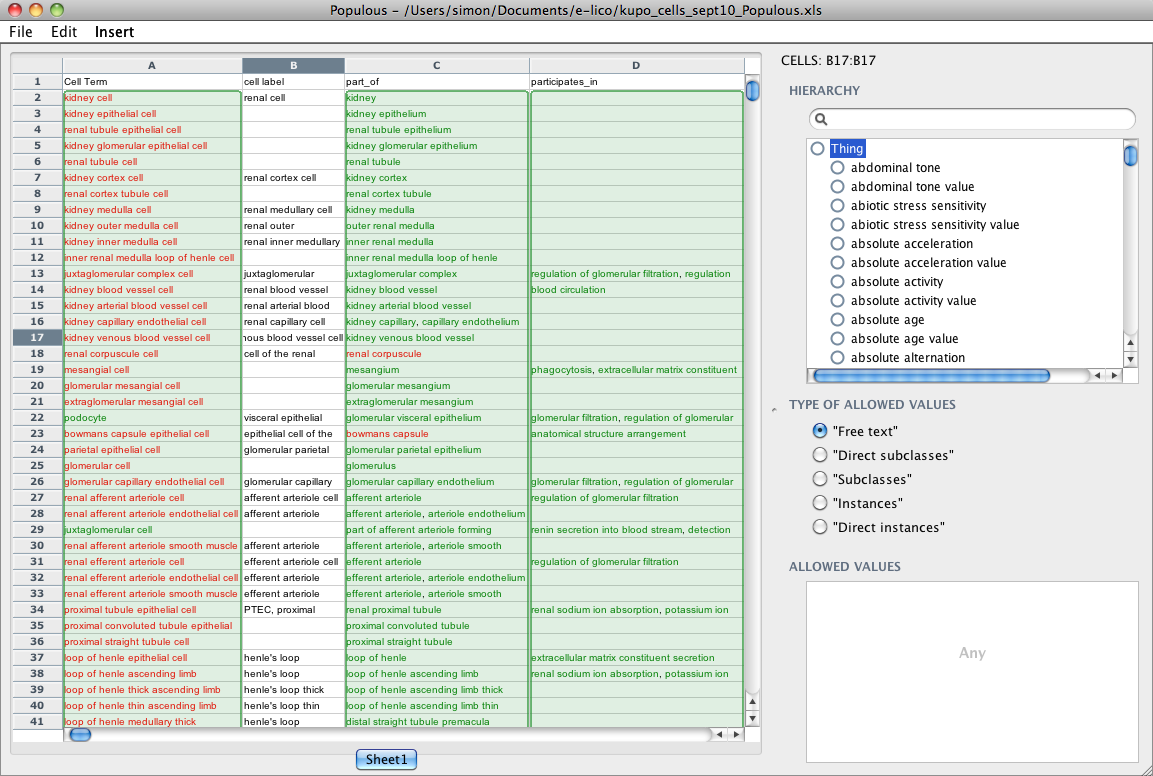}
\caption{\label{fig:pop_kupo}Screenshot of Populous showing template population for KUP ontology}
\end{figure}

A simple template was designed for experts from the KUP domain to capture the relationships between cell types, their anatomical location and their biological functions. The template has three main columns; column A is for entering cell type terms, column C is for anatomy terms and column D for biological process terms. \toolname was used to constrain the allowable values in columns A, C and D to concepts from the Open Biomedical Ontology Cell Type Ontology~\cite{cto}, subclasses or part of the \emph{Kidney} or \emph{Urinary system} concepts from the Mouse Adult Gross Anatomy Ontology~\cite{mao2005}, and all subclasses of the \emph{Biological Process} concept from the Gene Ontology~\cite{go2000}, respectively.  The experts were instructed that the relationship between concepts in column A and C was \emph{part of}, and the relationship between column A and D, \emph{participates in}. For concepts that were related to multiple concepts they were allowed to list concepts in a cell separated by a comma. Figure \ref{fig:pop_kupo} is a screen shot of \toolname populated with data from the domain experts. 

In order to transform the tabular data into an OWL representation the OPPL pattern in Figure \ref{fig:oppl_1} was created by the ontology engineers. This pattern states that a cell type is equivalent to a cell that is \partof an anatomy term and a subclass of cells that participate in biological processes.\footnote{Where possible we use the relationships from \protect\cite{Smith:2005p354}.}  For both restrictions the existential (\emph{some}) quantification is used.  The two differentia in this pattern for a cell genus are the anatomical location and the biological process, which is retrieved from column A, B and C respectively in the template.  The entire KUP ontology is generated from the template data combined with the ontology pattern. Figure \ref{fig:oppl_2} shows the Manchester OWL syntax generated from the pattern and data from row 13 for the \emph{Juxtaglomerular complex cell}.

\begin{figure}
\emph{Pattern 1}\\
\vocab{?cell:CLASS,}\\
\vocab{?anatomyPart:CLASS,}\\
\vocab{?partOfRestriction:CLASS = cell and part\_of some ?anatomyPart,}\\
\vocab{?anatomyIntersection:CLASS = createIntersection(?partOfRestriction.VALUES)}\\
\vocab{BEGIN}\\
\vocab{	ADD ?cell equivalentTo ?anatomyIntersection}\\
\vocab{END;}\\
\\
\emph{Pattern 2}\\
\vocab{?participant:CLASS,}\\
\vocab{?participatesRestriction:CLASS = ?cell and participates\_in some ?participant,}\\
\vocab{?participatesIntersection:CLASS = createIntersection(?participatesRestriction.VALUES)}\\
\vocab{BEGIN}\\
\vocab{	ADD ?cell SubClassOf ?participatesIntersection}\\
\vocab{END;}\\
\caption{\label{fig:oppl_1} Two OPPL 2 patterns for describing cell types in KUPO}
\end{figure}

\begin{figure}
\vocab{Class: kupo\_000027}\\
\vocab{   EquivalentTo: }\\
\vocab{        cell:CL\_0000000}\\
\vocab{        and (ro:part\_of some MA:MA\_0002580)}\\
\vocab{    SubClassOf: }\\
\vocab{        cell:CL\_0000000,}\\
\vocab{        ro:participates\_in some gene\_ontology:GO\_0002000,}\\
\vocab{        ro:participates\_in some gene\_ontology:GO\_0002001}\\
\caption{\label{fig:oppl_2} Manchester OWL syntax for Juxtaglomerular cell (MA\_0002580 = `part of afferent arteriole forming juxtaglomerular complex', GO\_0002000 = `detection of renal blood flow' and GO\_0002001 = `renin secretion into blood stream'}
\end{figure}

\begin{figure}
\centering
\includegraphics[height=7.5cm]{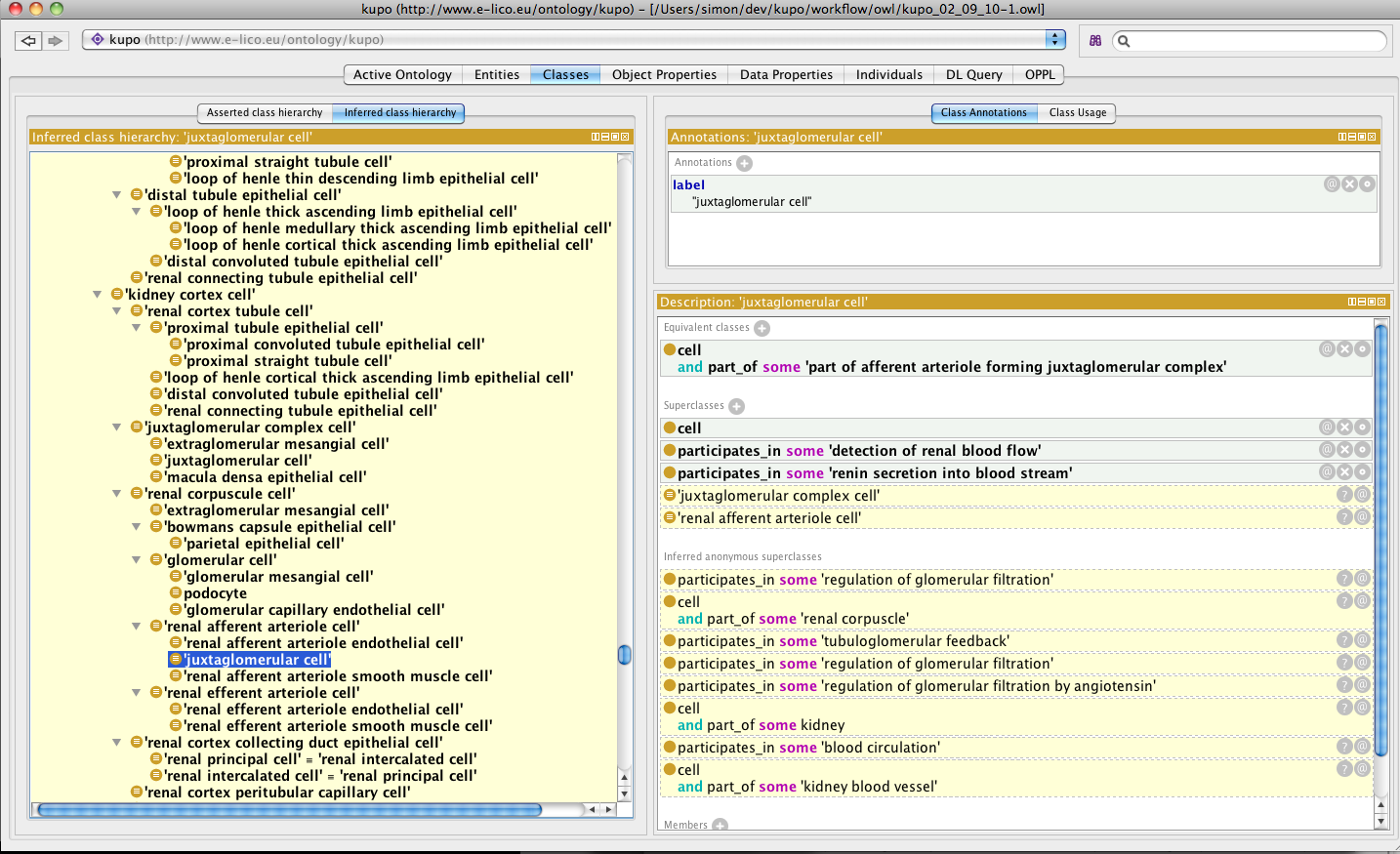}
\caption{\label{fig:kupo_protege}Screenshot of KUPO loaded into \protege 4.0 showing inferred class hierarchy for Juxtaglomerular cell}
\end{figure}

Using this template approach the domain experts described over 140 cell types, many of which are absent from the current CTO. Figure \ref{fig:kupo_protege} shows the inferred hierarchy after classifying the ontology in \protege 4.0. Note the asserted class hierarchy is simply a flat list of cell types, the partonomy of the mouse anatomy is used to drive inferences about super/sub class relationships between cell types. Leaving the reasoner to compute the class hierarchy allows the domain experts to inspect the ontology for missing or incorrect inferences. These often indicate some missing information in the template, or an error in the imported ontologies. This methodology was successful in engaging the domain experts to both contribute to the KUP ontology and generate new term requests for the imported ontologies.

\section{Discussion}

\toolname is designed for domain experts to gather knowledge that can be subsequently used to populate ontologies. Whilst previous tools have provided support for transforming templates into ontologies, they lacked basic support to help the user at the point of data entry. \toolname was designed to fill this niche and meet the requirements outlined in section \ref{sec:intro}.  The simple tabular interface used in \toolname is familiar to users who have already used a spreadsheet application. \toolname should lower the entry requirements for domain experts to contribute to ontology development projects. 

The release of \toolname as presented is an early version; there remains many possible additions, some of which are:
\begin{enumerate}
\item   \toolname can handle multiple values in a cell that maps to conjunctions of properties. Extensions to \toolname such that some minor syntax can be used to extend the  ability to use OWL's syntax would be useful. In particular, being able to specify numbers for cardinality and numbers  and other literals for  datatype properties.
\item   As already described, \toolname uses a row centric  model. We aim to use  \mtwo  to enable more variety in how tables are mapped to templates or patterns. For example, only portions of columns may be required to be mapped to certain axiom patterns and \mtwo enables this sort of mapping. OPPL and \mtwo together  should cover our mapping needs.
\item \toolname currently gathers domain knowledge for the ontology, but not about the ontology. We aim to extend \toolname to support various metadata such as editorial metadata and definitional metadata etc.
\item  \toolname is a single user application. Making \toolname collaborative such that contributors may collectively add material to the same spreadsheet.
\item Feedback from the generated ontology to fix or extend data in \toolname is currently \emph{ad hoc}. A tighter coupling of this feedback cycle, without having to go into an axiom based editor, will increase the quality assurance  aspects of \toolname.
 \end{enumerate}

We have demonstrated how \toolname can be used to develop an ontology describing cells of the kidney and urinary pathway system. This demonstration highlights how domain experts managed to generate a real application ontology without being exposed to an ontology language like OWL, or a tool like \protege. \toolname's main purpose is for knowledge gathering and not ontologising. By shielded users from the ontology, except for review later in the process, they are left to concentrate on the biology and not worry about the axioms needed to represent it. This separation is particularly useful should the ontologist wish to change the conceptualisation or experiment with different patterns for the representation.

Our experience in developing the KUPO with domain experts provided interesting insights into the benefits of developing an ontology in this way. Classical approaches to ontology development in the life science have tended to focus on building rich asserted hierarchies of concepts. The KUPO approach exploits the expressiveness of an ontology language like OWL to describe the cells in such a way that the class hierarchy is computed by the reasoner. This means we have a logical explanation for all the subsumptions in the hierarchy, that is useful for spotting erroneous or missing information. For example, there are cell types for the vasa recta descending limb and the vasa recta ascending limb, both of which have different functions. The imported anatomy ontology, however, only describes a vasa recta. The domain experts were able to spot this and can now submit a request for these two new concepts to be added to the anatomy ontology. Building normalised ontologies that facilitate the kinds of inferences we see in KUPO are generally considered to be harder and more time consuming than constructing class hierarchies manually, despite offering a clear benefit \cite{rector}. However, in cases where a repeating pattern can be abstracted from the ontology, as in the case of KUP cells, we see that domain experts can rapidly produce rich ontologies with considerably less investment using \toolname. 

The question now arises as to how far can you go with a tool like \toolname? \toolname is by no means a replacement for full blown ontology editors, nor is it intended to be. The scenarios where \toolname is of benefit assume that the ontology being developed has repeating patterns in the modelling. Furthermore, specifying the patterns for new ontologies before they exist is particularly difficult and is often something that emerges later as the ontology matures. For example, it was assumed with the kidney cells that we could describe them all in terms of their anatomy, only to later find some exceptions to the pattern. Renal principal and renal intercalated cells are currently indistinguishable by anatomy and function alone. There are always going to be exceptions, especially when modelling a complex domain like biology. We hope that \toolname can bring more domain experts into the ontology development process and engage them in the development process. 

The template approach can be particularly advantageous in scenarios where the modelling needs to change. \cite{qtt2009} showed how templates can be used to generate different ontological representations of the same data. The KUPO is also being used to link data in an RDF store to support the KUP KB, where only limited support for OWL inferences is possible. By developing a different set of patterns we could generate a simpler version of the KUP ontology from the same \toolname data for use in such an application. This again highlights an added benefit of separating the pattern from its population. 

OPPL provides an expressive language for generating OWL patterns. OPPL's support for variables make mapping single columns from tabular data to variables convenient.  The built in macros means we can create abstract pattern where the axioms can be generated dynamically depending on the number of values stored in a variable. \toolname is currently limited to working with uniform spreadsheets and assumes a row-per-entity paradigm, where single columns map to a particular variable. This structure keeps the template simple and should cover the majority of use cases for populating an ontology in this way. The extension to support multi-values per cell offers some additional flexibility over existing spreadsheet based approaches. In order to accommodate more complex spreadsheets we plan to extend \toolname to support more complex mappings from columns to spreadsheets. We are also exploring integrating the \mtwo language from MappingMaster directly into \toolname, we note that templates created in \toolname can already be exported as Excel document and loaded into MappingMaster for transformation should the user desire. 
  
\toolname offers a means of creating ontology content without the use of a standard ontology development tool. Just as data-entry tools exist for populating databases, so we also need such tools for populating ontologies. It is possible to separate knowledge gathering from conceptualisation and axiomatisation and \toolname is one means of achieving this goal. Such a separation offers flexibility  and simple form fill-in style of knowledge gathering that should make generation of axiomatically rich ontologies increasingly straight-forward.

\textbf{Acknowledgements:} We kindly acknowledge Mikel Ega\~na Aranguren for advice, requirements and testing \toolname. This work is funded by the e-LICO project---EU/FP7/ICT-2007.4.4 
and by SysMO-DB - BBSRC grant BBG0102181.

%
%
\bibliographystyle{plain}
\bibliography{references}

\end{document}